\documentclass[letterpaper]{article} 
\usepackage{aaai25}  
\usepackage{times}  
\usepackage{helvet}  
\usepackage{courier}  
\usepackage[hyphens]{url}  
\usepackage{graphicx} 
\usepackage{natbib}  
\usepackage{caption} 
\usepackage{algorithm}
\usepackage{algorithmic}
\usepackage{rotating}
\usepackage{multirow}
\usepackage{bigstrut}
\usepackage{amsmath}
\usepackage{amssymb}
\usepackage{xcolor}
\usepackage{booktabs}

\usepackage{newfloat}
\usepackage{listings}
\DeclareCaptionStyle{ruled}{labelfont=normalfont,labelsep=colon,strut=off} 
\floatstyle{ruled}
\newfloat{listing}{tb}{lst}{}
\floatname{listing}{Listing}
\nocopyright

\title{Adapting Segment Anything Model to Multi-modal Salient Object Detection with Semantic Feature Fusion Guidance}
\author{
    Kunpeng Wang\textsuperscript{\rm 1}, 
    Danying Lin\textsuperscript{\rm 1},
    Chenglong Li\textsuperscript{\rm 2}\thanks{Corresponding authors.},
    Zhengzheng Tu\textsuperscript{\rm 1}\footnotemark[1],
    Bin Luo\textsuperscript{\rm 1}
}
\affiliations{
    \textsuperscript{\rm 1}School of Computer Science and Technology, Anhui University, China\\
    \textsuperscript{\rm 2}School of Artificial Intelligence, Anhui University, China\\

    {kp.wang, danying\_lin, lcl1314}@foxmail.com, zhengzhengahu@163.com, luobin@ahu.edu.cn
}

\usepackage{bibentry}

\begin{document}

\maketitle

\begin{abstract}
Although most existing multi-modal salient object detection (SOD) methods demonstrate effectiveness through training models from scratch, the limited multi-modal data hinders these methods from reaching optimality. 
In this paper, we propose a novel framework to explore and exploit the powerful feature representation and zero-shot generalization ability of the pre-trained Segment Anything Model (SAM) for multi-modal SOD. 
Despite serving as a recent vision fundamental model, driving the class-agnostic SAM to comprehend and detect salient objects accurately is non-trivial, especially in challenging scenes. 
To this end, we develop \underline{SAM} with se\underline{m}antic f\underline{e}ature fu\underline{s}ion guidanc\underline{e} (Sammese), which incorporates multi-modal saliency-specific knowledge into SAM to adapt SAM to multi-modal SOD tasks. 
However, it is difficult for SAM trained on single-modal data to directly mine the complementary benefits of multi-modal inputs and comprehensively utilize them to achieve accurate saliency prediction.
To address these issues, we first design a multi-modal complementary fusion module to extract robust multi-modal semantic features by integrating information from visible and thermal or depth image pairs. Then, we feed the extracted multi-modal semantic features into both the SAM image encoder and mask decoder for fine-tuning and prompting, respectively. Specifically, in the image encoder, a multi-modal adapter is proposed to adapt the single-modal SAM to multi-modal information. In the mask decoder, a semantic-geometric prompt generation strategy is proposed to produce corresponding embeddings with various saliency cues. 
Extensive experiments on both RGB-D and RGB-T SOD benchmarks show the effectiveness of the proposed framework. 
The code will be available at \url{https://github.com/Angknpng/Sammese}.
\end{abstract}

\section{Introduction}
\label{sec:intro}

Salient object detection (SOD) aims to identify and segment the most visually attractive regions in visible images. It aids in reducing interfering information and has shown successful applications in various computer vision tasks, such as visual distraction reduction~\cite{aberman2022deep}, object recognition~\cite{flores2019saliency}, and image enhancement~\cite{miangoleh2023realistic}. Although promising progress has been achieved, it still struggles to handle challenging scenes, such as complex backgrounds and low illumination. With the availability of depth and thermal infrared sensors, some recent studies~\cite{qu2017rgbd,wang2018rgb} introduce depth maps or thermal images as an additional modality to complement the spatial or shape information of the corresponding visible images for high-quality saliency prediction, namely, RGB-Depth (RGB-D) and RGB-Thermal (RGB-T) SOD. 

\begin{figure}[t]
\centering
\includegraphics[width=1\columnwidth]{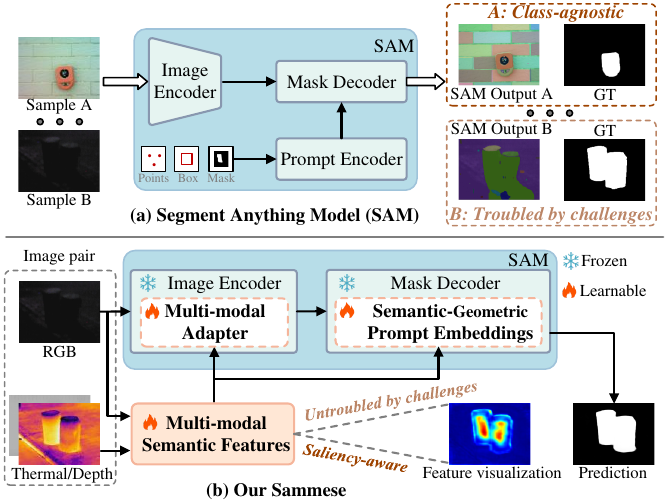}
\caption{A Comparison between SAM and our Sammese. Sammese incorporates multi-modal semantic features into SAM through the multi-modal adapter and semantic-geometric prompt embeddings, endowing it with the ability to be saliency-aware and untroubled by most challenges.}
\label{fig::motivation}
\end{figure}

Existing methods mainly integrate multi-modal information through different fusion strategies to improve the prediction results, such as early fusion, late fusion, and intermediate fusion~\cite{zhou2021rgb,pang2023caver}. They are almost trained from scratch, which requires a large amount of data to learn feature representations with saliency cues, especially for models with numerous parameters. However, due to the high acquisition and labeling costs, existing multi-modal datasets are scale-limited, which is not sufficient to support training an effective model in this manner. Although some methods utilize siamese networks~\cite{fu2021siamese}, few-shot learning~\cite{fu2022few}, and feature decoupling modules~\cite{chen2020rgbd} to alleviate data requirements, they are prone to overfitting current datasets and lack generalizability.

Recently, the Segment Anything Model (SAM)~\cite{kirillov2023segment} has gained significant attention as a fundamental vision segmentation model. Its core idea is to predict diverse and detailed segmentation masks based on user-provided prompts (i.e., points, boxes, and coarse masks), as shown in Fig.~\ref{fig::motivation} (a). Trained on massive masks and images, SAM demonstrates powerful zero-shot generalization and feature representation capabilities across a variety of vision tasks~\cite{yuan2024open,han2023boosting,wu2023medical}. Therefore, it is natural to introduce SAM into multi-modal SOD, leveraging its prior knowledge of segmentation to reduce the dependence on data. 
However, applying SAM to multi-modal SOD is non-trivial. On the one hand, as a class-agnostic model, SAM is unable to grasp object semantics, making it difficult to distinguish between salient and non-salient regions. On the other hand, SAM that is trained on single-modal images fails to achieve accurate segmentation in challenging scenes. The two samples in Fig.~\ref{fig::motivation} (a) illustrate each of these two issues. To address them, we incorporate multi-modal semantic features with rich saliency cues into SAM, endowing it with the ability to be saliency-aware and untroubled by most challenges, as shown in Fig.~\ref{fig::motivation} (b). 

\begin{figure}[t]
\centering
\includegraphics[width=1\columnwidth]{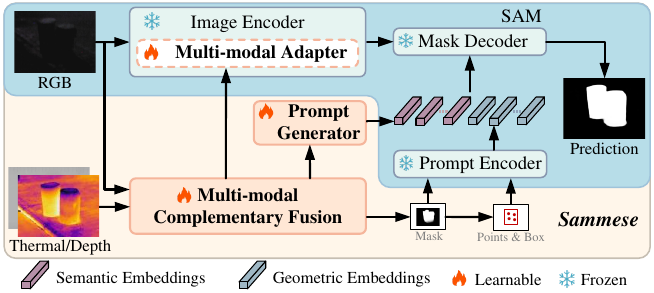}
\caption{An overview of the proposed Sammese.}
\label{fig::framework_easy}
\end{figure}

In this paper, we propose a novel multi-modal SOD framework called Sammese, which embeds saliency-specific knowledge into SAM with semantic feature fusion guidance. As illustrated in Fig.~\ref{fig::framework_easy}, in order to preserve the powerful segmentation ability of SAM, we freeze all its pre-trained parameters and learn additional parameters to fine-tune and prompt it, while the output by SAM in turn updates the learnable parameters. 
Specifically, in addition to the inherent visible image input of SAM, Sammese accepts the corresponding thermal or depth image as information supplement to deal with various challenges. However, since SAM is pre-trained on single-modal data, it is difficult for it to 1) directly mine the complementary benefits of multi-modal inputs and 2) comprehensively exploit the obtained multi-modal complementary information to achieve accurate saliency prediction. For the first issue, we design a multi-modal complementary fusion module to explore and exploit the complementary benefits between modalities. This module enables the extraction of effective multi-modal semantic feature representations, even in complex scenes. For the second issue, we propose a multi-modal adapter and a semantic-geometric prompt generation strategy to adequately incorporate the extracted multi-modal semantic features into SAM in different ways. The multi-modal adapter introduces the adaption technique~\cite{houlsby2019parameter} to adapt the single-modal SAM to multi-modal information. In particular, it appends a light yet effective fusion unit into the down-up bottleneck of the original adapter. With a lightweight design, the multi-modal adapter can be embedded into each block of SAM image encoder to compute adapted image embeddings and achieve multi-modal adaption. The semantic-geometric prompt generation strategy produces various prompt embeddings for the mask decoder to encourage SAM to segment out saliency-related regions. For the semantic prompt embeddings, we design a prompt generator that draws saliency cues from the multi-modal semantic features through a set of learnable queries. The geometric prompts that includes masks, boxes, and points are generated from the coarse saliency maps directly predicted by the multi-modal semantic features. They are fed into the SAM prompt encoder to produce the corresponding geometric prompt embeddings.
By sending the image and prompt embeddings into the mask decoder, the accurate saliency maps can be predicted by SAM.
Extensive experiments on both RGB-D and RGB-T SOD benchmarks show the effectiveness and superiority of the proposed framework.
The main contribution of our work can be summarized as follows.
\begin{itemize}
\item We propose a novel multi-modal SOD framework based on the pre-trained vision fundamental model SAM. Guided by multi-modal semantic features, SAM can be saliency-aware and untroubled by most challenges.
\item We design a multi-modal adapter, which is embedded into each SAM image encoder block with a simple yet effective structure to adapt SAM to multi-modal data.
\item We propose a semantic-geometric prompt generation strategy to produce various saliency-specific prompt embeddings for SAM without human intervention.
\item Our proposed Sammese achieves superior performance on six RGB-D and three RGB-T datasets, which inspires the application of foundation models in processing multi-modal data and performing multi-modal saliency detection tasks.
\end{itemize}


\begin{figure*}[t]
\centering
\includegraphics[width=1\textwidth]{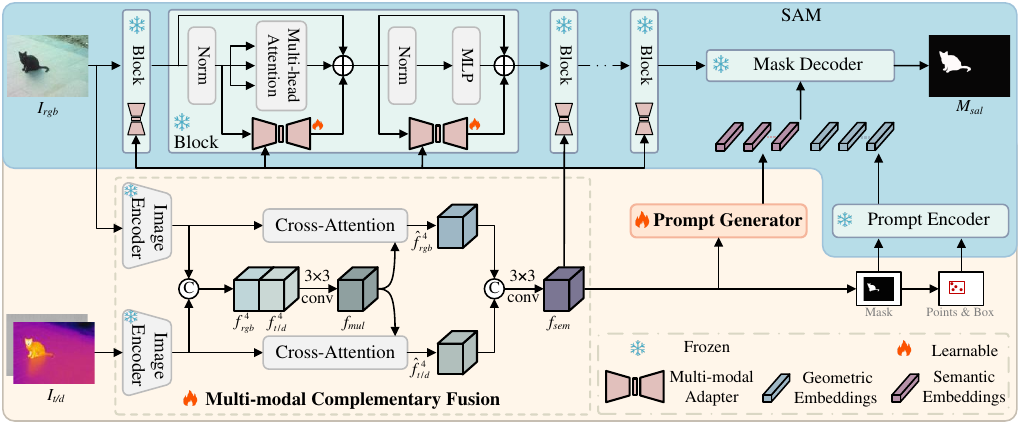}
\caption{Detailed architecture of the proposed Sammese framework. 
	The RGB-thermal or RGB-depth image pairs are fed into the multi-modal complementary fusion module to extract multi-modal semantic features, which are incorporated into SAM via multi-modal adapters and semantic-geometric prompt embeddings for multi-modal adaption and saliency prompting, respectively. With such comprehensive guidance, the parameter-frozen SAM is able to make accurate saliency predictions.}
\label{fig::framework}
\end{figure*}
\section{Related Work}
\label{sec:relatedWork}
\noindent\textbf{Multi-modal Salient Object Detection.}
Multi-modal Salient Object Detection (SOD) aims to improve the performance of RGB-based SOD by leveraging the complementary information of RGB-Depth (RGB-D) or RGB-Thermal (RGB-T) image pairs, which provide a comprehensive understanding of the scene. Conventional multi-modal SOD methods~\cite{ciptadi2013depth,peng2014rgbd} mainly utilize hand-crafted features extracted from image pairs and fuse them for prediction. Due to the limited expression ability of hand-crafted features, the performance of conventional multi-modal SOD is unsatisfactory. For improvement, recent studies employ deep neural networks to fuse multi-modal data. These methods can be roughly categorized into early fusion, late fusion and intermediate fusion in terms of fusion strategies~\cite{zhou2021rgb,pang2023caver}. 

Early fusion methods integrate visible images and depth or thermal images into a joint representation as input to a network. For example, Qu et al.~\cite{qu2017rgbd} compute saliency feature vectors from RGB-D image pairs and feed them into a convolutional neural network to extract representative unified RGB-D features. 
Late fusion methods mainly utilize the independent features or saliency maps from the two modalities to generate the final saliency prediction. Han et al.~\cite{han2018late} learn the feature representations of RGB and depth modalities separately, and mine their complementary relationships to obtain a joint representation for saliency prediction.
Intermediate fusion is widely used to effectively explore multi-modal correlations. For instance, Cong et al.~\cite{cong2023point} fuse multi-modal complementary features via point-aware interaction and CNN-induced refinement.
Although these methods improve the results of saliency prediction through different multi-modal fusion strategies, the limited multi-modal data hinders them from reaching optimality. Considering that the recent vision foundation segmentation model SAM is trained on large-scale data, we exploit its feature representation and generalization capabilities to accomplish multi-modal SOD.

\noindent\textbf{Application of SAM.}
Segmentation Anything Model (SAM) is a recently introduced vision foundation model pre-trained with 1 billion masks and 11 million images, emerging with impressive generalization and feature representation capabilities. It is a category-agnostic interactive model, which utilizes user instructions for segmentation, such as points, bounding boxes, and coarse masks. Currently, there are two main application approaches for SAM. One is to use the segmentation results of SAM as a prior to assist downstream tasks. For example, Han et al.~\cite{han2023boosting} integrate SAM with an open-vocabulary object detector to enhance its ability to detect arbitrary objects. Some studies~\cite{chen2023segment,xu2024multidimensional} utilize SAM to produce high-quality pseudo-labels for segmentation. The other is to improve the performance of SAM on downstream tasks through various prompts. SSOM~\cite{cui2023adaptive} uses original AdaLoRA~\cite{zhang2023adaptive} without modifications to fine-tune SAM for single-modal SOD. UV-SAM~\cite{zhang2024uv} utilizes a segmenter for SAM to generate mixed prompts for urban village identification.
These examples illustrate the effectiveness of SAM in different tasks. However, existing methods almost apply SAM to the original visible images, and they may fail in complex scenes. To break this limitation, we attempt to adapt SAM to multi-modal image pairs, undisturbed by most challenges. 

\section{Proposed Method}
\label{sec:method}
In this section, we describe the proposed novel multi-modal SOD framework Sammese, which drives the vision foundation model SAM to predict saliency maps with the guidance of multi-modal semantic features.

\noindent\textbf{Preliminary.} 
Given an aligned RGB and thermal or depth image pair $I_{rgb}$ and $I_{t/d}$, the goal of multi-modal SOD is to locate and segment out their common salient regions by learning a corresponding model $F_{sod}: \{ {I_{rgb}},{I_{t/d}}\}  \to {M_{sal}}$, where $M_{sal}$ is the predicted saliency map.

SAM is an interactive vision foundation model for class-agnostic segmentation. To be specific, SAM consists of an image encoder, a prompt encoder, and a mask decoder. The image encoder containing multiple transformer blocks is used to compute image embeddings. The prompt encoder produces prompt embeddings $P_{geo}$ from user-provided geometric prompts, including points, boxes or masks. The lightweight mask decoder integrates the image and prompt embeddings into a high-quality mask $M_{seg}$ related to the prompts. Accordingly, SAM can be formulated as ${F_{sam}}: \{ {I_{rgb}},{P_{geo}}\}  \to {M_{seg}}$.

\noindent\textbf{Overview.}
Fig.~\ref{fig::framework} illustrates the detailed architecture of the proposed Sammese framework, which takes RGB and thermal or depth image pairs as input. 
The multi-modal complementary fusion module integrates the image pairs to extract the saliency-specific multi-modal semantic feature ${f_{sem}}$, which is incorporated into SAM through multi-modal adapters and semantic-geometric embeddings. Guided by ${f_{sem}}$, SAM is able to make accurate saliency predictions. This process can be formulated as ${F_{sam}}:\{ {I_{rgb}},{f_{sem}}\}  \to {M_{sal}}$.
Saliency maps predicted from the multi-modal semantic feature and SAM are both supervised by ground truth. Additionally, all parameters of SAM are frozen to preserve its powerful segmentation ability.

\noindent\textbf{Multi-modal Complementary Fusion.}
The limited scale of multi-modal SOD datasets hinders the parameter learning of existing methods. Considering that SAM is pre-trained on large-scale data, we propose to transfer the segmentation capabilities of SAM to multi-modal SOD. However, the class-agnostic SAM is unable to grasp the semantics of salient objects and handle challenging scenes. Therefore, we design a Multi-modal Complementary Fusion Module (MCFM) to extract saliency-specific multi-modal semantic features.

\begin{figure}[t]
\centering
\includegraphics[width=1\columnwidth]{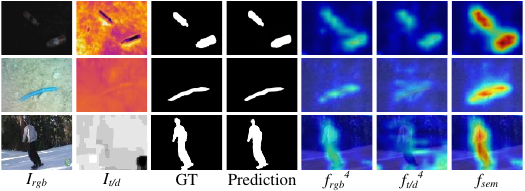}
\caption{Feature visualization for the proposed MCFM.}
\label{fig::visual}
\end{figure}

As shown in Fig.~\ref{fig::framework}, given the multi-modal image pairs $I_{rgb}$ and $I_{t/d}$, they are fed into a semantic-aware image encoder (i.e., Swin-B~\cite{liu2021swin}) to extract multi-level features $\{ f_{rgb}^i\} _{i = 1}^4$ and $\{ f_{t/d}^i\} _{i = 1}^4$. The image encoder is frozen during the training phase to prevent excessive learnable parameters and overfitting of MCFM. Since high-level features contain rich semantic information~\cite{zhang2018exfuse}, we select the top-level features of the two modalities (i.e., $f_{rgb}^4$ and $f_{t/d}^4$) for integration. Notably, the information richness of the two modalities varies with the scene. For example, thermal or depth images typically provide more object information in environments with complex backgrounds, whereas RGB images are superior in the case of low-quality thermal or depth images. Therefore, we concatenate RGB and thermal or depth features along the channel dimension and feed them into a convolutional layer to adaptively capture their relationships and integrate their different characteristics. This process can be formulated as:
\begin{equation}
\begin{split}
	&{f_{mul}} = Con{v_3}(concat(f_{rgb}^4,f_{t/d}^4)),
\end{split}
\end{equation} 
where $concat(,)$ represents the concatenation operation, $Con{v_x}$ denotes a convolution layer with kernel size of $x$, and $f_{mul}$ is the initial integrated multi-modal feature. Subsequently, we separately mine the information relevant to each modality from the integrated multi-modal feature through a cross-attention layer. This process can be summarized as:
\begin{equation}
\begin{split}
	&\hat f_m^4 = {\text{CrossAttn}}(f_m^4,{f_{mul}}),m = \{ rgb,t/d\} 
\end{split}
\end{equation} 
where $\hat f_{m}^4$ is the enhanced modality feature. Then, we employ a 3×3 convolutional layer to integrate the enhanced features of the two modalities:
\begin{equation}
\begin{split}
	&{f_{sem}} = Con{v_3}(concat(\hat f_{rgb}^4,\hat f_{t/d}^4)).
\end{split}
\end{equation} 
Ultimately, we obtain the integrated multi-modal semantic feature $f_{sem}$. 
Fig.~\ref{fig::visual} vividly demonstrates the visualization of features before and after MCFM. It can be seen that the multi-modal semantic feature extracted by MCFM is able to localize salient objects in various scenes, including low illumination (i.e., the first row), and low-quality thermal (i.e., the second row) and depth images (i.e., the third row).

\noindent\textbf{Multi-modal Adapter.}
Considering that SAM is trained on single-modal RGB images, we propose a Multi-modal Adapter (MAdapter) to adapt SAM to multi-modal data. MAdapter is inspired by the adaption technique~\cite{houlsby2019parameter} that is designed for parameter-efficient fine-tuning or transfer learning. 
Although recent work~\cite{cao2024bi} introduces the adaption technique to multi-modal scenarios, it directly employs the original adapter for cross-modal prompting and lacks sufficient multi-modal interactions. To this end, MAdapter incorporates multi-modal semantic guided fusion into the original adapter. With a lightweight design, MAdapter is embedded into the multi-head attention and MLP stages of each SAM image encoder block to ensure adequate adaptation, as illustrated among the SAM encoder of Fig.~\ref{fig::framework}.

\begin{figure}[t]
\centering
\includegraphics[width=1\columnwidth]{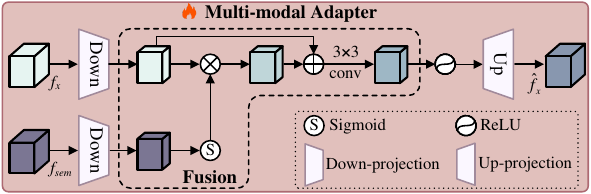}
\caption{The details of the proposed Multi-modal Adapter.}
\label{fig::adapter}
\end{figure}

To be specific, the original adapter consists of a down-projection linear layer followed by a nonlinear activation function and an up-projection linear layer. Given an intermediate input feature $f_{x} \in {\mathbb{R}^{d \times c}}$ at the $i$-th block, the original adapter can be formulated as:
\begin{equation}
\begin{split}
	&\text{Adapter}({f_x}) = \phi ({f_x}{W_{down}}){W_{up}},
\end{split}
\end{equation} 
where ${W_{down}} \in {\mathbb{R}^{c \times m}}$ is the down-projection layer, ${W_{up}} \in {\mathbb{R}^{m \times c}}$ is the up-projection layer, and $\phi (.)$ denotes the ReLU function. For specific details, please refer to the references~\cite{houlsby2019parameter,pfeiffer2021adapterfusion}.
As shown in Fig.~\ref{fig::adapter}, we introduce the multi-modal semantic feature and a corresponding fusion unit into the original adapter to form the MAdapter, which can be written as: 
\begin{equation}
\begin{split}
	&{\text{MAdapter}}({f_x},{f_{sem}}) = \phi ({F_{fus}}({f_x}{W_{down}},{f_{sem}}{W_{down}})){W_{up}},
\end{split}
\end{equation} 
where $F_{fus}$ refers to the fusion unit, which is operated in a compressed low-dimensional feature space to reduce computational complexity. In particular, we use a Sigmoid activation function to obtain the weight distribution of the low-dimensional multi-modal semantic feature, which is used to guide the low-dimensional intermediate input feature $f_x$ to highlight its salient regions. Then, the guided and original low-dimensional intermediate input feature are integrated through a residual connection and a 3×3 convolutional layer. This process can be formulated as:
\begin{equation}
\begin{split}
	&{F_{fus}} = Con{v_3}({f_x}{W_{down}} + {f_x}{W_{down}} \odot (\sigma ({f_{sem}}{W_{down}}))),
\end{split}
\end{equation} 
where $\sigma$ denotes the Sigmoid function, and $\odot$ represents the element-wise multiplication. 
In this way, the SAM image encoder is able to extract adapted image embeddings.

\noindent\textbf{Semantic-Geometric Prompt Generation.}
The prompts for SAM have to be provided interactively by humans, which increases labor costs and prevents the model from operating independently. To solve this issue, we propose a semantic-geometric prompt generation strategy that can directly produce corresponding prompt embeddings from the multi-modal semantic feature without human intervention. Notably, considering that the inherent geometric prompts of SAM only provide rough object location information, the strategy adds semantic prompt embeddings to further promote SAM to focus on salient regions.

\begin{figure}[t]
\centering
\includegraphics[width=1\columnwidth]{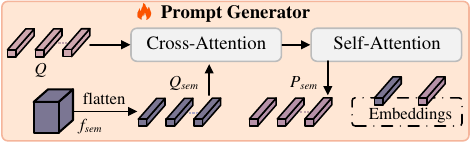}
\caption{The details of the proposed Prompt Generator.}
\label{fig::prompt}
\end{figure}

For the geometric prompt embeddings, we first feed the multi-modal semantic feature into a 1×1 convolutional layer and a upsampling layer to generate a coarse saliency map ${M_{sal\_coarse}}$ as the mask prompt, which can be described as: 
\begin{equation}
\begin{split}
	&{M_{sal\_coarse}} = Up(Con{v_1}({f_{sem}})),
\end{split}
\end{equation} 
where $Up(.)$ denotes the upsampling operation. Then, we perform image processing on the mask to derive the corresponding point and box prompts. Subsequently, the geometric prompt embeddings are obtained by feeding the prompts to the SAM prompt encoder.
For the semantic prompt embeddings, we design a Prompt Generator. As depicted in Fig.~\ref{fig::prompt}, we introduce a set of learnable queries $Q \in {\mathbb{R}^{N \times c}}$, where $N$ indicates the number of semantic prompts. Then, the queries engage with the flatten multi-modal semantic feature ${Q_{sem}} \in {\mathbb{R}^{hw \times c}}$ to obtain the saliency-specific information through the cross-attention and self-attention layer:
\begin{equation}
\begin{split}
	&P_{sem} = {\text{SelfAttn(CrossAttn(}}Q,{Q_{sem}}{\text{))}},
\end{split}
\end{equation} 
where $P_{sem}$ is the semantic prompt embeddings, possessing knowledge about the salient regions to be segmented. By feeding the geometric and semantic prompt embeddings and the adapted image embeddings into the mask decoder, the final saliency map $M_{sal}$ can be predicted.

\noindent\textbf{Loss Function.}
Following previous works~\cite{sun2024vrp,wu2021mobilesal}, we employ a combination of Binary Cross-Entropy (BCE) loss and Dice loss as the total loss ${\mathcal{L}_{total}}$: 
\begin{equation}
\begin{split}
	&{\mathcal{L}_{total}}(M,G) = {\text{BCE}}(M,G) + {\text{Dice}}(M,G),
\end{split}
\end{equation}
where $M$ is the saliency map, $G$ denotes the ground truth.
The total loss is used to supervise the saliency maps predicted by SAM and the MCFM. 

\section{Experiments}
\begin{table*}[htbp]
\centering
\resizebox{1\textwidth}{!}{
	\begin{tabular}{cc|cccccccccccc|c}
		\toprule[2.2pt]
		\multicolumn{2}{c|}{\multirow{3}[2]{*}{Method}} & DSA2F & DCF   & SPNet & VST   & MobileSal & CIR-Net & PICR-Net & LSNet & CAVER & LAFB  & VSCode & CPNet & Sammse \\
		\multicolumn{2}{c|}{} & CVPR21 & CVPR21 & ICCV21 & ICCV21 & TPAMI21 & TIP22 & ACM MM23 & TIP23 & TIP23 & TCSVT24 & CVPR24 & IJCV24 & - \\
		\multicolumn{2}{c|}{} &\cite{sun2021deep}       &\cite{Ji_2021_DCF}       &\cite{zhou2021specificity}       &\cite{liu2021visual}       &\cite{wu2021mobilesal}       &\cite{cong2022cir}       &\cite{cong2023point}       &\cite{zhou2023lsnet}       &\cite{pang2023caver}       &\cite{wang2024learning}       &\cite{luo2024vscode}       &\cite{hu2024cross}       & Ours \\
		\midrule
		\multicolumn{2}{c|}{Params (M) \hfill$\downarrow$} & 34.0  & 97.0  & 150.3  & 53.5  & 6.5   & 103.2  & 86.0  & \textbf{4.6}   & 55.8  & 118.8  & 74.7  & 247.8  & 68.0  \\
		\midrule
		\multirow{4}[2]{*}{\begin{sideways}\textit{DUT}\end{sideways}} & ${E_m}$ \hfill$\uparrow$   & 0.950  & 0.952  & 0.876  & 0.960  & 0.936  & 0.951  & 0.967  & 0.927  & 0.955  & 0.957  & 0.968  & 0.970  & \textbf{0.975} \\
		& ${S_m}$ \hfill$\uparrow$    & 0.921  & 0.924  & 0.803  & 0.943  & 0.896  & 0.932  & 0.943  & 0.886  & 0.931  & 0.931  & 0.952  & 0.951  & \textbf{0.954} \\
		& ${F_\beta}$ \hfill$\uparrow$   & 0.914  & 0.913  & 0.747  & 0.926  & 0.869  & 0.908  & 0.935  & 0.856  & 0.920  & 0.919  & 0.945  & 0.949  & \textbf{0.953} \\
		& $MAE$ \hfill$\downarrow$  & 0.030  & 0.030  & 0.085  & 0.025  & 0.044  & 0.031  & 0.022  & 0.049  & 0.029  & 0.027  & 0.019  & 0.019  & \textbf{0.016} \\
		\midrule
		\multirow{4}[2]{*}{\begin{sideways}\textit{NJUD}\end{sideways}} & ${E_m}$ \hfill$\uparrow$   & 0.923  & 0.922  & 0.931  & 0.913  & 0.914  & 0.922  & 0.930  & 0.891  & 0.922  & 0.924  & 0.930  & 0.935  & \textbf{0.938} \\
		& ${S_m}$ \hfill$\uparrow$    & 0.903  & 0.903  & 0.925  & 0.922  & 0.905  & 0.915  & 0.924  & 0.837  & 0.920  & 0.925  & \textbf{0.940} & 0.935  & 0.939  \\
		& ${F_\beta}$ \hfill$\uparrow$   & 0.889  & 0.884  & 0.909  & 0.892  & 0.874  & 0.881  & 0.909  & 0.775  & 0.903  & 0.910  & 0.925  & 0.926  & \textbf{0.934} \\
		& $MAE$ \hfill$\downarrow$  & 0.039  & 0.038  & 0.029  & 0.035  & 0.040  & 0.040  & 0.030  & 0.074  & 0.032  & 0.028  & 0.025  & 0.025  & \textbf{0.022} \\
		\midrule
		\multirow{4}[2]{*}{\begin{sideways}\textit{NLPR}\end{sideways}} & ${E_m}$ \hfill$\uparrow$   & 0.950  & 0.956  & 0.957  & 0.953  & 0.950  & 0.955  & 0.965  & 0.955  & 0.959  & 0.958  & 0.962  & 0.969  & \textbf{0.974} \\
		& ${S_m}$ \hfill$\uparrow$    & 0.918  & 0.921  & 0.928  & 0.931  & 0.920  & 0.933  & 0.935  & 0.918  & 0.929  & 0.929  & 0.938  & 0.941  & \textbf{0.942} \\
		& ${F_\beta}$ \hfill$\uparrow$   & 0.889  & 0.892  & 0.899  & 0.891  & 0.878  & 0.889  & 0.911  & 0.881  & 0.899  & 0.902  & 0.909  & 0.921  & \textbf{0.930} \\
		& $MAE$ \hfill$\downarrow$  & 0.024  & 0.023  & 0.021  & 0.024  & 0.025  & 0.023  & 0.019  & 0.024  & 0.022  & 0.021  & 0.020  & \textbf{0.016} & 0.017  \\
		\midrule
		\multirow{4}[2]{*}{\begin{sideways}\textit{SIP}\end{sideways}} & ${E_m}$ \hfill$\uparrow$   & 0.908  & 0.920  & 0.930  & 0.936  & 0.914  & 0.917  & 0.916  & 0.911  & 0.927  & 0.937  & 0.949  & 0.940  & \textbf{0.953} \\
		& ${S_m}$ \hfill$\uparrow$    & 0.861  & 0.873  & 0.894  & 0.903  & 0.873  & 0.888  & 0.865  & 0.909  & 0.893  & 0.897  & 0.917  & 0.908  & \textbf{0.919} \\
		& ${F_\beta}$ \hfill$\uparrow$   & 0.838  & 0.850  & 0.873  & 0.878  & 0.837  & 0.848  & 0.838  & 0.877  & 0.874  & 0.883  & \textbf{0.915} & 0.899  & 0.912  \\
		& $MAE$ \hfill$\downarrow$  & 0.057  & 0.051  & 0.043  & 0.040  & 0.054  & 0.053  & 0.056  & 0.040  & 0.043  & 0.041  & 0.032  & 0.036  & \textbf{0.028} \\
		\midrule
		\multirow{4}[2]{*}{\begin{sideways}\textit{SSD}\end{sideways}} & ${E_m}$ \hfill$\uparrow$   & 0.904  & 0.898  & 0.910  & 0.907  & 0.898  & 0.898  & 0.915  & 0.902  & 0.915  & 0.922  & 0.890  & 0.916  & \textbf{0.931} \\
		& ${S_m}$ \hfill$\uparrow$    & 0.876  & 0.852  & 0.871  & 0.889  & 0.863  & 0.862  & 0.878  & 0.856  & 0.874  & 0.881  & 0.862  & 0.894  & \textbf{0.901} \\
		& ${F_\beta}$ \hfill$\uparrow$   & 0.836  & 0.800  & 0.831  & 0.836  & 0.804  & 0.791  & 0.837  & 0.796  & 0.826  & 0.840  & 0.822  & 0.863  & \textbf{0.872} \\
		& $MAE$ \hfill$\downarrow$  & 0.047  & 0.053  & 0.044  & 0.045  & 0.052  & 0.054  & 0.046  & 0.055  & 0.044  & 0.041  & 0.049  & 0.035  & \textbf{0.028} \\
		\midrule
		\multirow{4}[2]{*}{\begin{sideways}\textit{STERE}\end{sideways}} & ${E_m}$ \hfill$\uparrow$   & 0.928  & 0.931  & 0.930  & 0.916  & 0.916  & 0.921  & 0.937  & 0.913  & 0.931  & 0.930  & 0.933  & 0.933  & \textbf{0.942} \\
		& ${S_m}$ \hfill$\uparrow$    & 0.897  & 0.905  & 0.907  & 0.913  & 0.903  & 0.891  & 0.920  & 0.871  & 0.914  & 0.906  & 0.928  & 0.921  & \textbf{0.929} \\
		& ${F_\beta}$ \hfill$\uparrow$   & 0.877  & 0.880  & 0.879  & 0.872  & 0.865  & 0.836  & 0.898  & 0.827  & 0.887  & 0.882  & 0.903  & 0.901  & \textbf{0.914} \\
		& $MAE$ \hfill$\downarrow$  & 0.038  & 0.037  & 0.037  & 0.038  & 0.041  & 0.049  & 0.031  & 0.054  & 0.034  & 0.037  & 0.030  & 0.029  & \textbf{0.026} \\
		\bottomrule[2pt]
	\end{tabular}%
}	
\caption{Quantitative comparisons with 12 state-of-the-art RGB-D SOD methods on six representative datasets. The best results are highlighted in \textbf{bold}. '$\uparrow$/$\downarrow$': a higher/lower score is better.}
\label{tab:RGBDCom}%
\end{table*}%

\label{sec:exp}
\subsection{Experiment Setup}
\textbf{Datasets and Evaluation Metrics.}
For RGB-D SOD, we conduct experiments on six prevalent datasets, including DUT~\cite{piao2019depth}, NJUD~\cite{ju2014depth}, NLPR~\cite{peng2014rgbd}, SSD~\cite{zhu2017three}, SIP~\cite{fan2020rethinking}, and STERE~\cite{niu2012leveraging}. Following recent works~\cite{cong2023point,sun2021deep}, we take a collection of 700 samples from NLPR, 1485 samples from NJUD and 800 samples from DUT to train our RGB-D SOD model.For RGB-T SOD, experiments are conducted on three widely used datasets, including VT821~\cite{tang2019rgbt}, VT1000~\cite{tu2019rgb}, VT5000~\cite{tu2020rgbt}. Following previous works~\cite{pang2023caver,tu2021multi}, we train our RGB-T SOD model on the training set of VT5000 with 2500 samples.
We employ four widely used evaluation metrics to assess performance, including S-measure (${S_m}$), E-measure (${E_m}$), F-measure (${F_\beta}$), and Mean Absolute Error ($MAE$). We also report the number of learnable parameters (Params) for complexity analysis.

\begin{table*}[htbp]
\centering
\resizebox{1\textwidth}{!}{
	\begin{tabular}{cc|cccccccccccc|c}
		\toprule[2pt]
		\multicolumn{2}{c|}{\multirow{3}[2]{*}{Method}} & MIDD  & CGFNet & SwinNet & DCNet & TNet  & HRTransNet & OSRNet & LSNet & CAVER & WaveNet & LAFB  & VSCode & Sammse \\
		\multicolumn{2}{c|}{} & TIP21 & TCSVT21 & TCSVT21 & TIP22 & TMM22 & TCSVT22 & TIM22 & TIP23 & TIP23 & TIP23 & TCSVT24 & CVPR24 & - \\
		\multicolumn{2}{c|}{} &\cite{tu2021multi}       &\cite{wang2021cgfnet}       &\cite{liu2021swinnet}       &\cite{tu2022weakly}       &\cite{cong2022does}       &\cite{tang2022hrtransnet}       &\cite{huo2022real}       &\cite{zhou2023lsnet}       &\cite{pang2023caver}       &\cite{zhou2023wavenet}       &\cite{wang2024learning}       &\cite{luo2024vscode}       & Ours \\
		\midrule
		\multicolumn{2}{c|}{Params (M) \hfill$\downarrow$} & 52.4  & 66.4  & 199.2  & 24.1  & 87.0  & 26.3  & 15.6  & \textbf{4.6} & 55.8  & 80.7  & 118.8  & 74.7  & 68.0  \\
		\midrule
		\multirow{4}[2]{*}{\begin{sideways}\textit{VT821}\end{sideways}} & ${E_m}$ \hfill$\uparrow$   & 0.895  & 0.912  & 0.926  & 0.912  & 0.919  & 0.929  & 0.896  & 0.911  & 0.919  & 0.929  & 0.915  & \textbf{0.945} & 0.943  \\
		& ${S_m}$ \hfill$\uparrow$    & 0.871  & 0.881  & 0.904  & 0.876  & 0.899  & 0.906  & 0.875  & 0.878  & 0.891  & 0.912  & 0.884  & 0.920  & \textbf{0.924} \\
		& ${F_\beta}$ \hfill$\uparrow$   & 0.760  & 0.829  & 0.818  & 0.823  & 0.841  & 0.849  & 0.801  & 0.809  & 0.835  & 0.863  & 0.817  & 0.882  & \textbf{0.887} \\
		& $MAE$ \hfill$\downarrow$  & 0.045  & 0.038  & 0.030  & 0.033  & 0.030  & 0.026  & 0.043  & 0.033  & 0.033  & 0.024  & 0.034  & 0.021  & \textbf{0.020} \\
		\midrule
		\multirow{4}[2]{*}{\begin{sideways}\textit{VT1000}\end{sideways}} & ${E_m}$ \hfill$\uparrow$   & 0.933  & 0.944  & 0.947  & 0.948  & 0.937  & 0.945  & 0.935  & 0.935  & 0.945  & 0.952  & 0.945  & \textbf{0.958} & 0.957  \\
		& ${S_m}$ \hfill$\uparrow$    & 0.915  & 0.923  & 0.938  & 0.922  & 0.929  & 0.938  & 0.926  & 0.925  & 0.936  & 0.945  & 0.932  & \textbf{0.949} & 0.947  \\
		& ${F_\beta}$ \hfill$\uparrow$   & 0.856  & 0.900  & 0.894  & 0.902  & 0.895  & 0.913  & 0.891  & 0.887  & 0.909  & 0.921  & 0.905  & 0.931  & \textbf{0.934} \\
		& $MAE$ \hfill$\downarrow$  & 0.027  & 0.023  & 0.018  & 0.021  & 0.021  & 0.017  & 0.022  & 0.023  & 0.017  & 0.015  & 0.018  & 0.014  & \textbf{0.013} \\
		\midrule
		\multirow{4}[2]{*}{\begin{sideways}\textit{VT5000}\end{sideways}} & ${E_m}$ \hfill$\uparrow$   & 0.897  & 0.922  & 0.942  & 0.920  & 0.927  & 0.945  & 0.908  & 0.915  & 0.924  & 0.940  & 0.931  & 0.946  & \textbf{0.955} \\
		& ${S_m}$ \hfill$\uparrow$    & 0.868  & 0.883  & 0.912  & 0.871  & 0.895  & 0.912  & 0.875  & 0.877  & 0.892  & 0.911  & 0.893  & 0.918  & \textbf{0.919} \\
		& ${F_\beta}$ \hfill$\uparrow$   & 0.763  & 0.831  & 0.846  & 0.819  & 0.840  & 0.870  & 0.807  & 0.806  & 0.835  & 0.864  & 0.841  & 0.880  & \textbf{0.889} \\
		& $MAE$ \hfill$\downarrow$  & 0.043  & 0.035  & 0.026  & 0.035  & 0.033  & 0.025  & 0.040  & 0.037  & 0.032  & 0.026  & 0.030  & 0.025  & \textbf{0.022} \\
		\bottomrule[2pt]
	\end{tabular}%
}
\caption{Quantitative comparisons with 12 state-of-the-art RGB-T SOD methods on three representative datasets.}
\label{tab:RGBTCom}%
\end{table*}%
\noindent\textbf{Implementation Details.}
The proposed framework is implemented using the PyTorch toolbox on an A100 GPU. The pre-trained SAM is equipped with a ViT-B backbone, and the image encoder uses the pre-trained Swin-B. The input images for SAM and MCFM are resized 1024×1024 and 384×384, respectively. The number of learned queries is set to 30 by default.
During training, we adopt the AdamW algorithm for optimization. The learning rate is set as 1e-5. The models for both tasks converge within 100 epochs with the batch size of 2.

\subsection{Quantitative Comparisons}
We compare our method with 12 state-of-the-art (SOTA) RGB-D methods and 12 SOTA RGB-T SOD methods, respectively.
The quantitative results on six RGB-D and three RGB-T SOD datasets are shown in Table~\ref{tab:RGBDCom} and Table~\ref{tab:RGBTCom}. It can be seen that our method overall outperforms all compared methods across the nine multi-modal SOD datasets. 
To be specific, compared with the sub-optimal RGB-D SOD method (i.e., CPNet) , our method achieves average improvements of 0.9\%, 0.6\%, 1.0\% , and 14.4\% on the four evaluation metrics (i.e., $E_m$, $S_m$, ${F_\beta}$, and $MAE$) for the six datasets, respectively. In the RGB-T SOD comparison, compared to the on the advanced method VSCode, our method achieves an average improvement of 3.9\% across the four evaluation metrics on the most challenging VT5000 dataset. This validates that the proposed framework effectively adapts SAM to multi-modal SOD tasks.


\subsection{Ablation Studies}

\begin{table}[t]
\centering
\resizebox{1\columnwidth}{!}{
	\begin{tabular}{c|c|c|ccc|ccc|ccc}
		\toprule[1.5pt]
		\multirow{2}[4]{*}{ID} & \multirow{2}[4]{*}{Model} & \multirow{2}[4]{*}{Params} & \multicolumn{3}{c|}{DUT} & \multicolumn{3}{c|}{STERE} & \multicolumn{3}{c}{VT5000} \bigstrut\\
		\cmidrule{4-12}          &       &       & ${S_m}$ \hfill$\uparrow$   & ${F_\beta}$ \hfill$\uparrow$   & $MAE$ \hfill$\downarrow$  & ${S_m}$ \hfill$\uparrow$   & ${F_\beta}$ \hfill$\uparrow$   & $MAE$ \hfill$\downarrow$  & ${S_m}$ \hfill$\uparrow$   & ${F_\beta}$ \hfill$\uparrow$   & $MAE$ \hfill$\downarrow$ \bigstrut\\
		\midrule
		0 & Sammse & 68.0  & \textbf{0.954} & \textbf{0.953} & \textbf{0.016} & \textbf{0.929} & \textbf{0.914} & \textbf{0.026} & \textbf{0.919} & \textbf{0.889} & \textbf{0.022} \bigstrut\\
		\midrule
		1     & w/o MCFM & 32.3 & 0.935  & 0.930  & 0.024  & 0.911  & 0.890  & 0.034  & 0.903  & 0.867  & 0.030  \bigstrut[t]\\
		2     & w/ CD & 101.6  & 0.952  & 0.947  & 0.017  & 0.923  & 0.907  & 0.029  & 0.917  & 0.885  & 0.023  \bigstrut[b]\\
		\bottomrule[1.25pt]
	\end{tabular}%
}
\caption{Ablation analyses of the multi-modal complementary fusion module (MCFM) 'CD': complex design.}
\label{tab:aba_mcfm}%
\end{table}%

\begin{figure}[t]
\centering
\includegraphics[width=1\columnwidth]{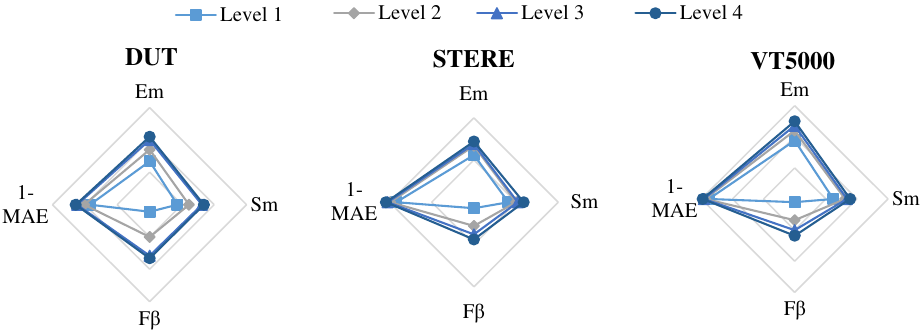}
\caption{Ablation study on MCFM with different layers.}
\label{fig::layer}
\end{figure}
\noindent\textbf{Effect of MCFM.}
In order to verify the effectiveness of the proposed MCFM, we directly remove it, denoted as 'w/o MCFM' in Table~\ref{tab:aba_mcfm}. This means that the complementary information of the two modalities cannot be fully exploited. Specifically, we concatenate the top-level features of the two modalities by channel dimension, and then feed them into a 3×3 convolution layer for simple fusion. Compared to our full model on the three datasets, the three metrics drops on average by 1.9\%, 2.6\%, and 39.1\%, respectively. We also replace MCFM with a more complex design, denoted as 'w/ CD' in Table~\ref{tab:aba_mcfm}. Specifically, the complex design utilizes a self-attention after both the fused features $f_{mul}$ and $f_{sem}$ in MCFM to further interact the multi-modal fused features. The comparison (ID2 $vs.$ ID1, ID2 $vs.$ ID0) shows that the complex design brings performance gains, but inferior to our MCFM. This is mainly because complex designs with a large number of parameters (i.e., 101.6 MB) can easily lead to overfitting. 
In addition, we employ features from different levels of the image encoder for MCFM to observe their impact on the performance, as shown in Fig.~\ref{fig::layer}. As the level becomes shallower, the semantic information decreases, making it difficult for MCFM to grasp sufficient semantics of salient objects, leading to performance decrease.

\begin{figure}[t]
\centering
\includegraphics[width=1\columnwidth]{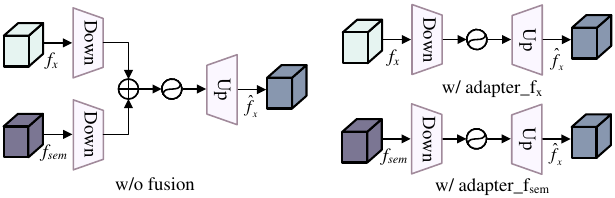}
\caption{Structures for the ablation study of MAdapter.}
\label{fig::aba_madpter}
\end{figure}

\noindent\textbf{Effect of MAdapter.}
To verify the validity of MAdapter, we directly remove it, noted as 'w/o MAdapter' in Table~\ref{tab:aba_MAdapter}. This means that it is difficult for SAM to adapt from single-modal to multi-modal data. The comparison between ID1 and ID0 proves the positive effect and lightweight design of MAdapter. To investigate the effectiveness of the MAdapter in detail, we further conduct three variants: remove the fusion unit of the MAdapter (i.e., 'w/o fusion'), use the original adapter with only the intermediate feature of SAM encoder block (i.e., 'w/ Adapter\_f$_x$'), and use the original adapter with only the multi-modal semantic feature (i.e., 'w/ Adapter\_f$_{sem}$'). The corresponding structures are shown in Fig.~\ref{fig::aba_madpter}. As reported in Table~\ref{tab:aba_MAdapter}, the performance degradation of 'w/o fusion' (e.g., ${S_m}:0.954 \to 0.947$, ${F_\beta}:0.953 \to 0.943$, $MAE:0.016 \to 0.019$) validates that the fusion unit benefits to the multi-modal adaption for SAM. The performance of 'w/ Adapter\_f$_x$' also decreases (e.g., ${S_m}:0.954 \to 0.946$, ${F_\beta}:0.953 \to 0.941$, $MAE:0.016 \to 0.020$), indicating that the adapter without multi-modal semantic guidance fails to adequately drive SAM for accurate saliency prediction. In addition, the relatively slight performance degradation of 'w/ Adapter\_f$_{sem}$' (${S_m}:0.929 \to 0.926$, ${F_\beta}:0.914 \to 0.910$, $MAE:0.026 \to 0.027$) suggests that although the multi-modal semantic features are effective, they need to be integrated with the intermediate features to achieve better multi-modal adaptation. Furthermore, by comparing ID2 and ID4, it can be found that direct summation hardly achieves effective integration of multi-modal information in the low dimension.
\begin{table}[t]
\centering
\resizebox{1\columnwidth}{!}{
	\begin{tabular}{c|c|c|ccc|ccc|ccc}
		\toprule[1.5pt]
		\multirow{2}[4]{*}{ID} & \multirow{2}[4]{*}{Model} & \multirow{2}[4]{*}{Params} & \multicolumn{3}{c|}{DUT} & \multicolumn{3}{c|}{STERE} & \multicolumn{3}{c}{VT5000} \bigstrut\\
		\cmidrule{4-12}          &       &       & ${S_m}$ \hfill$\uparrow$   & ${F_\beta}$ \hfill$\uparrow$   & $MAE$ \hfill$\downarrow$  & ${S_m}$ \hfill$\uparrow$   & ${F_\beta}$ \hfill$\uparrow$   & $MAE$ \hfill$\downarrow$  & ${S_m}$ \hfill$\uparrow$   & ${F_\beta}$ \hfill$\uparrow$   & $MAE$ \hfill$\downarrow$ \bigstrut\\
		\midrule
		0 & Sammse & 68.0  & \textbf{0.954} & \textbf{0.953} & \textbf{0.016} & \textbf{0.929} & \textbf{0.914} & \textbf{0.026} & \textbf{0.919} & \textbf{0.889} & \textbf{0.022} \bigstrut\\
		\midrule
		1     & w/o MAdapter & 51.1 & 0.943  & 0.937  & 0.021  & 0.918  & 0.898  & 0.031  & 0.909  & 0.875  & 0.028  \bigstrut[t]\\
		2     & w/o Fusion & 58.2  & 0.947  & 0.943  & 0.019  & 0.924  & 0.912  & 0.028  & 0.915  & 0.881  & 0.024  \\
		3     & w/ Adapter\_f$_x$ & 58.2  & 0.946  & 0.941  & 0.020  & 0.921  & 0.904  & 0.030  & 0.912  & 0.879  & 0.026  \\
		4     & w/ Adapter\_f$_{sem}$ & 58.2  & 0.949  & 0.947  & 0.019  & 0.926  & 0.910  & 0.027  & 0.914  & 0.883  & 0.024  \bigstrut[b]\\
		\bottomrule[1.5pt]
	\end{tabular}%
}
\caption{Ablation analyses of the proposed MAdapter.}
\label{tab:aba_MAdapter}%
\end{table}%

\begin{figure}[t]
\centering
\includegraphics[width=1\columnwidth]{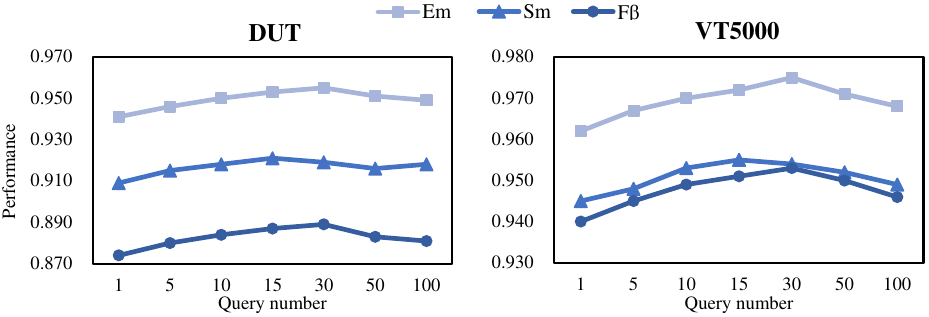}
\caption{Ablation study for different query numbers.}
\label{fig::query}
\end{figure}

\noindent\textbf{Effect of the semantic-geometric prompt generation strategy.}
To verify the effectiveness of the semantic and geometric prompts, we remove them successively, denoted as 'w/o Semantic' and 'w/o Geometric' in Table~\ref{tab:aba_propmt}, respectively. By comparing ID1 and ID0, it can be found that the semantic prompt improves the three metrics on the three datasets, especially achieves the percentage gain of 25\% for $MAE$ score on the DUT dataset. In addition, without the geometric prompt, the three evaluation
metrics drop by an average of 1.6\%, 2.3\%, and 34.4\%, respectively.
In Fig~\ref{fig::query}, we also conduct a comprehensive analysis of the impact of varying semantic query quantities on the performance of Sammese. We observe a positive correlation between increased query numbers and improved segmentation quality. However, once the number of queries exceeds 30, the performance gains begin to diminish. This phenomenon suggests that the semantic prompt of 30 queries provides sufficient guidance to the model, and further increases in the query count do not yield significant improvements.
\begin{table}[t]
\centering
\resizebox{1\columnwidth}{!}{
	\begin{tabular}{c|c|c|ccc|ccc|ccc}
		\toprule[1.5pt]
		\multirow{2}[4]{*}{ID} & \multirow{2}[4]{*}{Model} & \multirow{2}[4]{*}{Params} & \multicolumn{3}{c|}{DUT} & \multicolumn{3}{c|}{STERE} & \multicolumn{3}{c}{VT5000} \bigstrut\\
		\cmidrule{4-12}          &       &       & ${S_m}$ \hfill$\uparrow$   & ${F_\beta}$ \hfill$\uparrow$   & $MAE$ \hfill$\downarrow$  & ${S_m}$ \hfill$\uparrow$   & ${F_\beta}$ \hfill$\uparrow$   & $MAE$ \hfill$\downarrow$  & ${S_m}$ \hfill$\uparrow$   & ${F_\beta}$ \hfill$\uparrow$   & $MAE$ \hfill$\downarrow$ \bigstrut\\
		\midrule
		0 & Sammse & 68.0  & \textbf{0.954} & \textbf{0.953} & \textbf{0.016} & \textbf{0.929} & \textbf{0.914} & \textbf{0.026} & \textbf{0.919} & \textbf{0.889} & \textbf{0.022} \bigstrut\\
		\midrule
		1     & w/o Semantic & 57.5 & 0.945  & 0.940  & 0.020  & 0.916  & 0.901  & 0.031  & 0.911  & 0.878  & 0.026  \bigstrut[t]\\
		2     & w/o Geometric & 68.0  & 0.939  & 0.931  & 0.022  & 0.911  & 0.892  & 0.036  & 0.907  & 0.871  & 0.028  \bigstrut[b]\\
		\bottomrule[1.5pt]
	\end{tabular}%
}
\caption{Ablation analyses of the proposed semantic-geometric prompt generation strategy.}
\label{tab:aba_propmt}%
\end{table}%
\begin{table}[h!]
\centering
\resizebox{1\columnwidth}{!}{
	\begin{tabular}{c|c|c|ccc|ccc|ccc}
		\toprule[1.5pt]
		\multirow{2}[4]{*}{ID} & \multirow{2}[4]{*}{Model} & \multirow{2}[4]{*}{Params} & \multicolumn{3}{c|}{DUT} & \multicolumn{3}{c|}{STERE} & \multicolumn{3}{c}{VT5000} \bigstrut\\
		\cmidrule{4-12}          &       &       & ${S_m}$ \hfill$\uparrow$   & ${F_\beta}$ \hfill$\uparrow$   & $MAE$ \hfill$\downarrow$  & ${S_m}$ \hfill$\uparrow$   & ${F_\beta}$ \hfill$\uparrow$   & $MAE$ \hfill$\downarrow$  & ${S_m}$ \hfill$\uparrow$   & ${F_\beta}$ \hfill$\uparrow$   & $MAE$ \hfill$\downarrow$ \bigstrut\\
		\midrule
		0     & Sammse & 68.0  & \textbf{0.954} & \textbf{0.953} & \textbf{0.016} & 0.929  & \textbf{0.914} & \textbf{0.026} & \textbf{0.919} & \textbf{0.889} & \textbf{0.022} \bigstrut\\
		\midrule
		1     & VSCode~\cite{luo2024vscode} & 74.7  & \textbf{0.954} & 0.949  & 0.019  & \textbf{0.930} & 0.908  & 0.029  & \textbf{0.919} & 0.883  & 0.024  \bigstrut[t]\\
		2     & CAVER~\cite{pang2023caver} & 55.8  & 0.934  & 0.923  & 0.028  & 0.916  & 0.891  & 0.033  & 0.895  & 0.836  & 0.032  \\
		3     & LSNet~\cite{zhou2023lsnet} & 4.6   & 0.891  & 0.866  & 0.045  & 0.875  & 0.834  & 0.051  & 0.881  & 0.811  & 0.036  \bigstrut[b]\\
		\midrule
		4     & SAM w/ points & 0.0   & 0.889  & 0.878  & 0.048  & 0.886  & 0.835  & 0.051  & 0.846  & 0.730  & 0.049  \bigstrut[t]\\
		5     & SAM w/ adapter & 16.9  & 0.920  & 0.911  & 0.032  & 0.899  & 0.875  & 0.040  & 0.864  & 0.785  & 0.043  \bigstrut[b]\\
		\bottomrule[1.5pt]
	\end{tabular}%
}
\caption{Ablation analyses of SAM.}
\label{tab:samEffect}%
\end{table}%

\noindent\textbf{Effect of SAM.}
To reduce the dependence on limited multi-modal SOD data, our model introduces the ViT-B backbone of SAM pre-trained on SA-1B. For fairness, we also introduce the ViT-B backbone into some advanced models by feature summation without extra changes. The results of ID1-ID3 in Table~\ref{tab:samEffect} show improvements, but not significant and overall inferior to our method. In addition, we validate the baseline SAM. ID4 shows that the performance of SAM under point prompts sampled from encoder features is comparable to some specific methods, and ID5 shows that SAM fine-tuned by our MAdapter can be further improved.

\section{Conclusion}
\label{sec:conclusion}
In this paper, we propose Sammese, a novel multi-modal SOD framework based on the recent vision foundation model SAM. The Sammese framework introduces multi-modal semantic features into SAM to drive it to comprehend and segment salient objects in various challenging scenes. To this end, we first design a multi-modal complementary fusion module to integrate saliency-specific multi-modal semantic features, which are then incorporated into SAM through two elaborated components (i.e., the multi-modal adapter and semantic-geometric prompts). The multi-modal adapter is embedded into each SAM image encoder block to adapt it to multi-modal information. The semantic-geometric prompts with various saliency cues are generated to encourage SAM to segment saliency-related regions. Experimental results demonstrate that our Sammese achieves superior performance on both RGB-D and RGB-T datasets, demonstrating a significant leap in leveraging the vision foundation model to process multi-modal data and conduct multi-modal saliency detection tasks.

\bibliography{aaai25_sammese}

\end{document}